# FOLKTALENT: ENHANCING CLASSIFICATION AND TAGGING OF INDIAN FOLK PAINTINGS

**Nancy Hada[1], Aditya Singh[2], Kavita Vemuri[3]**

[1,2]Department of Computer Science, IIIT Hyderabad, India
[3]Department of Cognitive Science, IIIT Hyderabad, India

**Abstract**
*Indian folk paintings have a rich mosaic of symbols, colors, textures, and stories making them an invaluable repository of cultural legacy. The paper presents a novel approach to classifying these paintings into distinct art forms and tagging them with their unique salient features. A custom dataset named FolkTalent, comprising 2279 digital images of paintings across 12 different forms, has been prepared using websites that are direct outlets of Indian folk paintings. Tags covering a wide range of attributes like color, theme, artistic style, and patterns are generated using GPT4, and verified by an expert for each painting. Classification is performed employing the RandomForest ensemble technique on fine-tuned Convolutional Neural Network (CNN) models to classify Indian folk paintings, achieving an accuracy of 91.83%. Tagging is accomplished via the prominent fine-tuned CNN-based backbones with a custom classifier attached to its top to perform multi-label image classification. The generated tags offer a deeper insight into the painting, enabling an enhanced search experience based on theme and visual attributes. The proposed hybrid model sets a new benchmark in folk painting classification and tagging, significantly contributing to cataloging India's folk-art heritage.*

## 1. INTRODUCTION

India showcases an incredible diversity of folk cultures, each characterized by geographical uniqueness. The diversity is profoundly reflected in the field of art [1], where each community presents a unique style of folk paintings depicting their day-to-day activities like agriculture, marriage, religious practices, hunting, etc. These paintings are valuable because of their distinctive use of colors, symbols, and textures, which serve as historical records of the respective folk lifestyle in addition to being an artistic expression. This historical significance makes it important to preserve them in their original form and protect their geographical tags. With this motivation, the proposed work is carried out to efficiently classify *twelve types* of folk paintings, namely *Warli*, *Bhil*, *Gond*, *Kalighat*, *Kalamkari*, *Pichwai*, *Rogan*, *Mata Ni Pachedi*, *Madhubani*, *Tanjore*, *Pattachitra,* and *Phad*. In addition to classification, the proposed tag generation mechanism can help in offering an enhanced search experience to the customers.

Each folk art is unique in its narrative, representation of objects/animals/humans, and the colors or material used. *Warli* art from Maharashtra stands out for its depiction of social life. Geometric patterns like triangles (mountains) and circles (sun) are used to depict harmony with nature. *Gond* art comes from central India and is characterized by its vibrant colors as well as its use of dashes, lines, and dots. *Bhil* art, also from central India, captures the folklore and their daily lifestyle using only colorful dots and motifs. *Kalamkari* art from Andhra Pradesh and Telangana portrays narratives from Hindu mythology. These can be found on temple walls and on cotton textiles. *Madhubani* art from Bihar is characterized by its complex geometric patterns, mostly depicting religious themes. *Pichwai* art originates from Rajasthan and is celebrated for its detailed depictions of Lord Krishna (a popular Hindu god). Lotus and cow are the main elements in *Pichwai* paintings. Gujarat's *Mata Ni Pachedi* depicts cultural stories representing the Mother Goddess. *Rogan* art is another art form practiced in Gujarat in which, thick castor-oil-based paints are used to create fluid dark-colored designs. *Tanjore* paintings from Tamil Nadu can be identified by the gold foil work giving them a luminous appearance. *Phad* paintings from Rajasthan mostly depict deities and local heroes through scrolls that narrate epic tales. *Pattachitra* from Odisha is mainly known for its intricacy depicting Hindu mythology on large cloth-based scrolls. The *Kalighat* paintings from West Bengal, originally souvenirs from the Kalighat temple, have evolved into a distinct style featuring bold brushstrokes and reflecting societal themes. An artistic representation of the Indian folk arts can be found in footnote[1].

Classification of paintings has been well explored within the field of computer vision [2]-[5]. Convolutional Neural Networks (CNNs) [8] serves as the fundamental framework for various downstream tasks like classification [7], [8], object detection [9], [10], image segmentation [11], [12] etc. Four prominent CNN models namely, ResNet-50 [13], EfficientNet-B0 [14], Inception-V3 [15] and VGG-16 [16] have made significant contributions in the field of image classification [2]-[7] and tagging [17]-[19]. Based on these findings, the CNN models seem to learn and perform efficiently on image classification tasks.

The foundational VGG model [16], introduced in 2014, highlighted simplicity through a deep stack of 3x3 convolutional layers, serving as robust baseline. Inception model or GoogLeNet [15] includes parallel filters and an inception module to capture features at different scales. ResNet [13] brought a revolution in the field with its introduction in 2015, mainly for its *skip* connection trick that helped it overcome vanishing gradient problem. EfficientNet [14] optimized both, computational cost as well as model size, balancing depth, width, and resolution through the principle of compound scaling. These models represent the primary milestones in the evolution of CNNs and continue to offer diverse solutions in the field of deep learning for computer vision applications.

---
[1]https://cdn.shopify.com/s/files/1/1194/1498/files/Folk_Art_Map_of_India_2019.jpg

With this background, the motivation behind the proposed work is classifying and tagging Indian folk paintings using CNN based architectures in multi-class and multi-label classification setting, respectively. To further fine-tune the performance and utilizing the popular CNN models collectively to make classification decision, RandomForest [20] based ensemble technique is used to aggregate the performance by each strong performing CNN model to arrive at a collective decision.

As part of the proposed work, the main contribution includes:
(1) The collation of a dataset, *FolkTalent*, that consists of 12 types of folk paintings with an additional comprehensive array of tags describing colors, attributes, objects, patterns, theme, and their unique elements of saliency.
(2) A novel approach of utilizing fine-tuned CNN backbones with RandomForest as an aggregator to efficiently classify Indian folk paintings into 12 classes improving overall accuracy significantly as compared to the existing works [21]-[25].
(3) The generation of associated tags for each image using the prominent fine-tuned backbones, to offers a deeper insight into the granular details in the painting.

The next section describes the existing literatures of folk paintings datasets, past works on classifications and tagging of Indian folk paintings, followed by the motivation behind the proposed work.

## 2. LITERATURE SURVEY

There have been multiple attempts [21]-[25] to create a dataset for Indian folk arts. This section covers a comprehensive overview of existing datasets, followed by an analysis of the prior works on Indian folk arts, thereby setting the ground for the motivation behind the proposed method.

**Datasets**: Varshney *et al.* [21] proposed a dataset comprising of five different forms of Madhubani paintings namely "Bharni, Godna, Kachni, Kohbar, and Tantrik". It contains 680 digital images collected from websites as used in ours. This dataset focuses only on one main art form. Kumar *et al.* [22] compiled a dataset of over 2400 paintings from 8 different folk arts namely *Mural*, *Pattachitra*, *Kalamkari*, *Portrait*, *Madhubani*, *Warli*, *Kangra*, and *Tanjore*. This was the first successful attempt at creating a dataset containing a diverse range of folk painting styles, with equal distribution across the art types. A similar attempt was made by Mane and Shrawankar [23] with around 1000 paintings across 26 different folk-art forms. The dataset size is too small to cover the huge diversity of 26 classes. A recent attempt was made to create a dataset [26] for Indian Visual arts, which includes four broad categories of artforms, namely, Sculpture, Pottery, Painting and Architecture. It consists of 4000 images, 1000 from each class, sourced from Google images. Further, Podder *et al.* [24] created a dataset focusing exclusively on Indian monuments which contains images of various sculptures and paintings depicting mythological stories. Images are taken from different angles and under different illuminations.

Numerous studies have been conducted using various tasks on these datasets to retrieve desired patterns. As the paintings are analyzed, it is crucial to capture and analyze visual patterns to draw conclusions for the hypothesis. With the evolution of CNN based models it becomes lot more efficient to address the nature of proposed work, that is image classification via transfer learning on Indian folk arts.

**Existing classification approaches**: A recent study [21] explored the classification of five forms of Madhubani paintings via transfer learning with a pre-trained CNN. This work is divided into two modules, classification using 'handcrafted features' and 'automatically extracted features.' The second approach leverages CNN-based backbones like, InceptionV3 [15] and InceptionResNet [27] for feature extraction and reported a classification accuracy of 98.82%. Despite having a high classification accuracy, the scalability of this dataset is limited due to its confined focus to just one folk artform. Another attempt [22] was made to classify Indian paintings using Support Vector Machine (SVM) [28] on the features extracted from CNN models like AlexNet [59] and VGG [16]. The proposed dataset was perfectly balanced, consisting of 8 different art forms on which they obtain a classification accuracy of 86.56%. Another work demonstrates an efficient query-based retrieval [23], where an accuracy of 82% was obtained on a dataset of 1000 paintings with 26 classes, via transfer learning (on pre-trained CNN) approach. Having few samples overall per class leads to the memorization (overfitting) issue thereby limiting its generalization and wide-scale application.

To summarize, the main challenges in the existing works include (1) limited diversity in the dataset, (2) a scope of improving classification score for a wider spectrum of Indian folk paintings, (3) skewed category-wise distribution within the datasets, and (4) the absence of validation split while training CNN models. Additionally, the existing literature focuses only on high level details of paintings and does not delve into the granular details including unique elements of saliency inherent in the paintings.

To address the gaps in the existing works, a new classification method is proposed involving an ensemble-based technique applied on extracted features from multiple fine-tuned CNN backbones. Our dataset boasts a larger volume of 2279 images across 12 diverse forms of folk paintings. Furthermore, the proposed work is the first attempt on image tagging on Indian folk paintings and aims at devising a model to efficiently generate tags for 12 different forms of folk paintings, which can be used for efficient retrieval using keywords, in future.

The next section includes a detailed discussion about the proposed dataset followed by the method used for performing multi-class and multi-label classification for image classification and tagging respectively.

## 3. PROPOSED WORK

This section is aimed at discussing the proposed dataset and the methodology. Section 3.1 talks about the proposed dataset, *FolkTalent* that covers 12 different forms of Indian folk paintings and shows their compositionality per class with split proportion for training, validation, and testing. To capture better granularity, GPT4 [29] was utilized to generate tags for each image. Section 3.2 elaborates on classification and tagging methodology adopted on *FolkTalent*. The Section 3.3 highlights the ensemble approach performed across fine-tuned CNN models.

### 3.1 *FolkTalent* Dataset

The *FolkTalent* dataset comprises 2279 images across 12 different classes of folk paintings. The images are collected from the websites [30]-[48] which are the direct sellers of Indian folk paintings. For each painting, the bio of the painter has been scrutinized to ensure the authenticity of the paintings. Each painter whose work is posted on sites has inherited the artform as an accomplished artist. This was one of the main criteria for choosing images for the dataset to ensure no counterfeits are considered for the analysis.

The dataset contains paintings of *Bhil* (191 images), *Gond* (183 images), *Mata Ni Pachedi* (185 images), *Kalighat* (184 images), *Kalamkari* (184 images), *Madhubani* (187 images), *Pattachitra* (195 images), *Phad* (214 images), *Pichwai* (187 images), *Tanjore* (191 images), *Rogan* (185 images) and *Warli* (190 images). Additionally, for each painting, an array of tags/keywords are generated using GPT4 [29] followed by a manual review. The tags include colors used, theme, patterns used (like dots, dashes, etc.), art style, and the elements of saliency like flora, fauna, celestial elements (sun, moon, and stars), human figures, agricultural elements, deities, etc. Figure 2 is an example of a painting from *FolkTalent* with its tags listed in footnote[2]. On average, 30 tags[2] are generated for each painting.

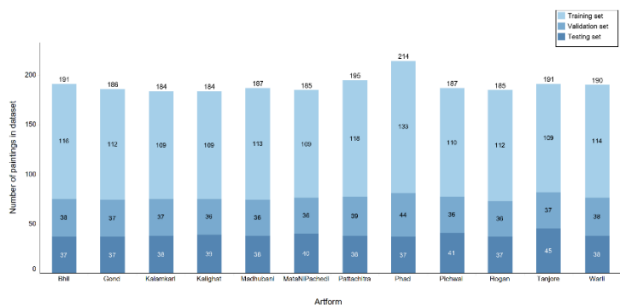

Fig.1. Image distribution across classes and partitions

The dataset is split into 3 sets, training, validation, and testing to assess the reliability and generalizability of proposed models. Training split containing 1364 images (60%), is used to iteratively adjust the model parameters with the goal of minimizing the loss function. Whereas to maintain the model's generalizability, the model's performance was consistently checked on the validation set which comprises of 450 images, that is 20% of the dataset (Figure 1) in every pass. Dataset balance is maintained to ensure similar set size per class.

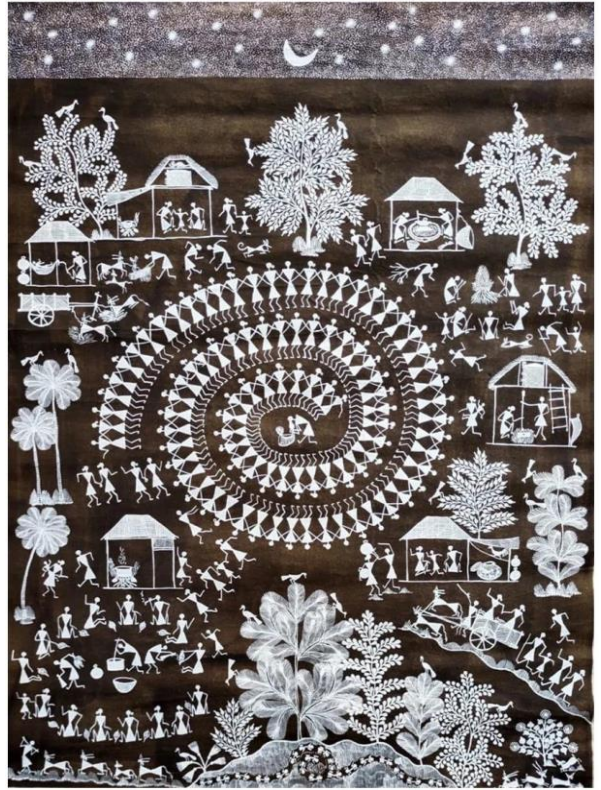

Source: A Warli Village, Warli Art by Dilip Bahotha, Image link

Fig. 2: A sample painting from the dataset along with the tags

The validation set is essential for assessing model's performance during training, enabling fine-tuning, and preventing overfitting by evaluating the model on unseen data. Incidentally, validation split has not been used so far in the existing works of classification of Indian folk paintings. Models are typically checkpointed based on the best validation score, raising concerns about a potential overfitting to the validation set. To ensure an unbiased evaluation of the model's performance on unseen data, the test dataset is introduced. In proposed work the test set comprises of 20% of images. This set mimics the real-world scenario by exposing the model to completely unseen data. Notably, all the three partitions are mutually exclusive and exhaustive.

**Pre-processing:** It is a critical initial step in any machine learning or data analysis work, aimed at preparing raw data for further analysis or modeling which involves a series of operations to clean, transform, and organize the data to make it suitable for the specific task at hand. The process begins with the frame and the background removal from the painting to avoid conflict with the actual content in the painting. The border within the painting was kept intact as it contributes to saliency and has unique significance based on the folk community and the themes. For example, *Warli* paintings depicting a wedding scene mostly contain a floral border as its hallmark feature. For tagging, the captioning potential of GPT4 was leveraged to generate keywords and tags, describing the key aspects of the painting. Semantically similar words were aggregated and "synonym replacement" was done to represent them via a single tag, e.g., GPT4 generated tags like 'celebrated', 'celebrating', 'feast', 'celebration', 'festivity', were all represented with a single keyword "celebration". Next, a word vocabulary of Next, a word vocabulary of 1500 tags, is generated consisting of all such tags generated.

---

[2] **Tags:** "Warli_art", "dark_brown_background", "stars", "moon", "white_figures", "dancing", "cooking", "cows", "birds", "fields", "white_patterns", "playing_instruments", "animal_herding", "riding_cart", "floral_patterns", "crops", "ploughing", "huts", "village_houses", "canopies", "musical_instruments", "celebration", "harvesting", "dance", "geometric_designs", "horses", "trees", "circular_dance_pattern", "circular_dance", "group_activities"

Using this vocabulary, a multi-label encoder transformed labels (tags) into a binary vector to indicate the presence of each tag in the vocabulary. This prepares the data for multi-label classification task. A demonstration of the tag-to-binary vector transformation is provided in Tables 1 and 2.

Table 1: Image tags for various samples

| Image | Tags |
|---|---|
| Image 1 | Sun, Stars, Cow |
| Image 2 | Lotus, Cow |
| Image 3 | Stars, Lizard |
| Image 4 | Sun, Stars |

Table 2: Binary vectors for four sample images

| Image | Sun | Stars | Lotus | Cow | Lizard |
|---|---|---|---|---|---|
| Image 1 | 1 | 1 | 0 | 1 | 0 |
| Image 2 | 0 | 0 | 1 | 1 | 0 |
| Image 3 | 0 | 1 | 0 | 0 | 1 |
| Image 4 | 1 | 1 | 0 | 0 | 0 |

**Data Augmentation:** As the name suggests, it is used to expand the dataset by applying various transformations and modifications to the existing dataset thus increasing its diversity and size. As variation invokes regularization, this in turn improves robustness of the model. Unlike usual image augmentation, color-based augmentation has been skipped in the proposed method, to preserve the significance of various colors in each artform and maintain their authenticity. Further, transformations like horizonal and vertical flips, as well as scaling that are standard practice to apply, have been carried out. Finally, all the images were resized to 224x224 pixels across three channels, RGB (Red Green Blue).

The final dataset following pre-processing, and augmentation comprises 2279 images, each sized at 224x224 pixels. Additionally, it includes binary vectors associated with tagging information for each image.

### 3.2 Fine-tuning CNN models for classification

**Problem Statement:** Given a folk painting image as input, the objective is to accurately classify it into one of twelve predefined categories, rendering a multi-class classification task. Further, to enhance our understanding of the image content, the associated tags (unique salient elements specific to each category) are predicted for each image, thereby making it a multi-label classification task. The generated tags serve to provide interpretability and insight into the model's decision-making process.

**Transfer learning with pre-trained CNN & their pre-processor:** With the proposed dataset, the task is to build an algorithm which efficiently classifies Indian folk paintings. Based on the promising performance on ImageNet classification dataset [58], four pre-trained CNN architectures, namely, VGG16, ResNet50, EfficientNetB0 and InceptionV3 were employed. These four, being unique in their modular architecture, offer distinct perspectives for analyzing images. For example, ResNet50 captures detailed features at different levels of abstraction due to its *skip* connections allowing it to handle complex patterns effectively, whereas EfficientNetB0 maintains efficiency in terms of model size and computational cost to compute features, while InceptionV3 leverages multi-scale feature extraction using parallel convolutional pathways. This enables the model to capture both local and global features effectively. Combining these viewpoints helps the model make better predictions and become more robust and generalizable. The CNN backbones mentioned require the images in a specific format due to the way they are trained. Hence, all the images are resized to 224x224 pixel resolution (except for InceptionV3 with 299x299) across all 3 RGB channels. Further, they are normalized across channels using corresponding pre-computed statistics from ImageNet [58].

**Fine Tuning CNN:** In the proposed analysis, the model is trained with a batch size of 128 for 100 epochs. Early stopping with a patience of 15 is applied to stop training the if no further improvement is observed in validation accuracy for 15 consecutive epochs. The model is checkpointed to save the last best weights based on validation accuracy. Additionally, learning rate is optimized using "Reduce Learning rate on Plateau (ReduceLROnPlateau)" [49] to enable maximum learning and a reduction in loss. ReduceLROnPlateau [49] effectively handles slow convergence by overcoming plateaus in the loss function surface. To achieve this, the patience was set to 8, which specifies the LR controller to drop the learning rate (reduction factor) to 50% of current if validation accuracy (monitoring criteria) does not improve for 8 consecutive epochs. The final set of hyper-parameter values is finalized based on best validation accuracy.

**Model Optimization:** For optimization, Adam (Adaptive moment estimation) [50] is used that combines the merits of two popular optimizers called RMSprop [51] and Momentum [52] for effectively updating the model's weight during training. Overall, it adapts learning rate to increase optimization with accumulated gradients from the previous passes that helps in smoother updates. The convergence observed is much faster than traditional algorithms like Stochastic Gradient Descent (SGD) [53] especially in case of varying curvature irregular loss [50]. As loss function, "Categorical cross-entropy (CE)" was used which is most preferable in multi-class classification tasks. It computes dissimilarity (entropy) between the predicted probability scores (per-class probability) and the true label (one-hot vector). With the objective of minimizing this loss function, models try to assign a higher probability score for correct class and lower for the incorrect ones. In the case of tagging, "Binary cross-entropy (BCE)" loss function is used, since it involves multi-label classification. The working principle of BCE is similar to CE, with the only difference being, all the classes in multi-label setting are considered to be independent.

**Classification head**: A basic structure of CNN architecture can be seen in Fig. 3. This is a module (head) of simple feed-forward network (FFN) [57] comprising of 12 neurons. It is added atop of the *headless* pre-trained CNN backbones to tailor the entire architecture to the specific task *i.e.,* classifying 12 classes instead of the default 1000 [27]. The backbone (excluding the FFN) acts as a feature descriptor and thus *unfreezed* during training to finetune (adapt) on proposed dataset. Parallelly, FFN (the classifier) too gets trained to correctly identify the labels. Similarly, for tagging, this FFN architecture is adapted to accommodate different number of classes (tags). In the design, FFN is added after Global Average Pooling (GAP) layer which produces 1024-dimensional features capturing the overall semantics of an image. These features serve as an input for the FFN to generate desired labels. Architecturally, FFN comprises a fully connected *linear* layer with 1024 output neurons followed by, ReLU [54] (Rectified Linear Unit) activation function, and another *linear* layer with 12 output neurons, followed by *Softmax* activation [55] to derive class probabilities. For tagging, the last *linear* layer is modified to have 1500 output neurons with *Sigmoid* activation function [56]. Unlike *Softmax*, *Sigmoid* treats all the labels independently.

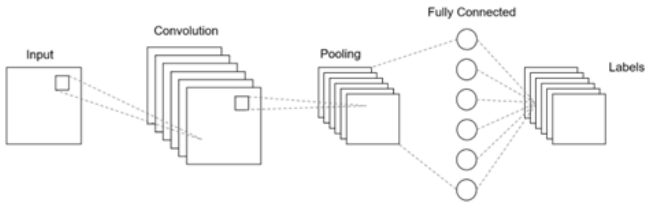

Fig. 3. CNN basic architecture [44, 53]

**Inference:** In this stage, the entire pipeline explained previously is executed, starting from the image-processing, and continuing until obtaining predictions from the classification head, all while maintaining the model's weights *frozen*. For multi-class classification, the class index is determined by taking the *argmax* of the predicted vector, which is then looked up against a *class-index* dictionary to obtain the corresponding class name. For tagging, the soft prediction scores are binarized using a threshold of 0.5 and all *tag* names corresponding to the vector elements that exceed the threshold are selected.

### 3.3 Ensembling Fined-tuned CNN architectures with Random Forest

Ensembling offers a powerful approach to harnessing the collective decisions of various models, enhancing overall robustness and reducing bias. Here, a similar strategy was adopted by incorporating a Random Forest [25] (a decision-tree-based ensembling technique) into the workflow. The probabilistic predictions from the top three fine-tuned CNN models (ResNet50, Inception V3, and EfficientNetB0) were combined and fed into the Random Forest algorithm as shown in Fig. 4. This helped in generating a more informed decision regarding the correct label by leveraging the diverse insights offered by each model.

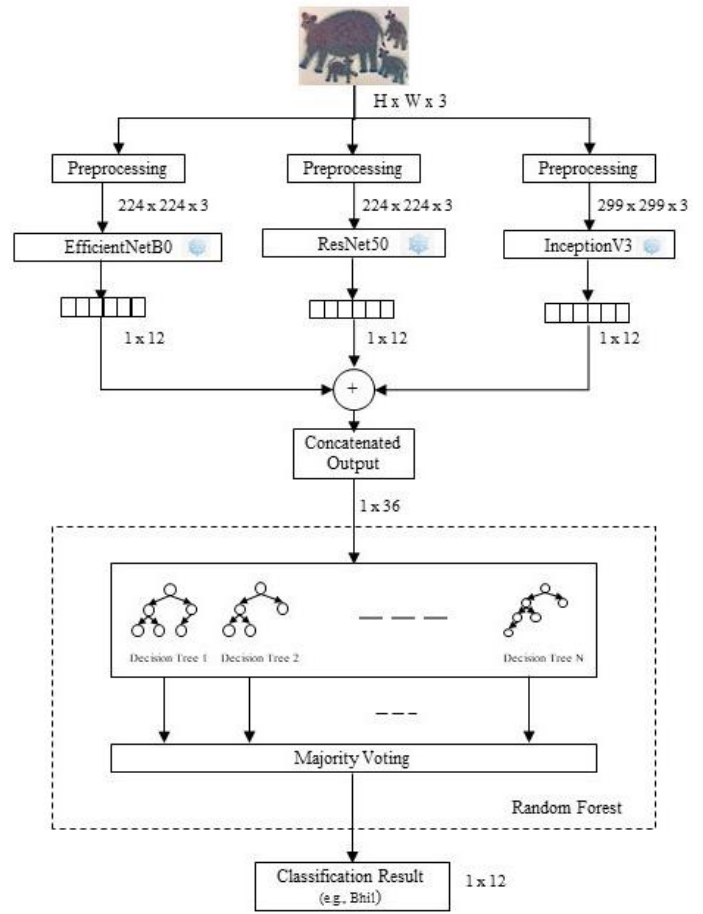

Fig. 4: Ensemble of fine-tuned CNN models using RandomForest

## 4. RESULTS

In this section, the results are presented that are obtained from CNN fine-tuning as well as ensembling as shown in Table 3.

Table 3: Performance of fine-tuned CNN models and their ensemble on FolkTalent for multi-class classification

| CNN Model | Training Accuracy | Validation Accuracy | Testing Accuracy |
|---|---|---|---|
| VGG16 | 75.28 | 73.83 | 62.8 |
| InceptionV3 | 99.93 | 90.44 | 88.82 |
| ResNet50 | 99.78 | *94* | 90.53 |
| EfficientNetB0 | 99.93 | 91.61 | 90.97 |
| Random Forest (Ensemble) | **100** | **94.00** | **91.83** |

As evident from the table, EfficientNetB0 has achieved the highest test accuracy, scoring 90.97%, while ResNet50 achieved highest validation accuracy of 94% with test accuracy being 90.53%. Further, VGG16 exhibited the lowest test accuracy of 62.8% as well as a low training and validation accuracy of 75.28% and 73.83%, respectively. One of the possible reasons is the *vanishing gradient* problem due to lack of skip-connections which ResNet has incorporated. This limits its ability to focus on intricate details which might be the defining feature of the image. Also, unlike pre-training of VGG, the dataset possesses a lot of elements (objects) leading to lot more confusion. Additionally, VGG lacks the ability to see images at multiple scales, unlike Inception. These aspects reflect its inability towards efficient classification for the use case and hence, is not considered in further analysis. InceptionV3 on contrary showed better performance with an accuracy of 88.82% on test and 90.44% on validation split. Next, we combine (ensemble) decisions from Inception, ResNet, and EfficientNet via Random Forest to achieve the best score on both validation and test, *i.e.,* 94% and 91.83%, respectively.

To have a detailed understanding of class-wise results, confusion matrix is generated for each CNN counterpart as well as for ensemble. For ResNet50, EffcientNetB0 and InceptionV3, confusion matrices are shown in figures 3, 4, and 5, respectively. Each matrix shows a significant classification score with minimal errors. Further, Table 4. shows the combined results from all four methods including ensemble.

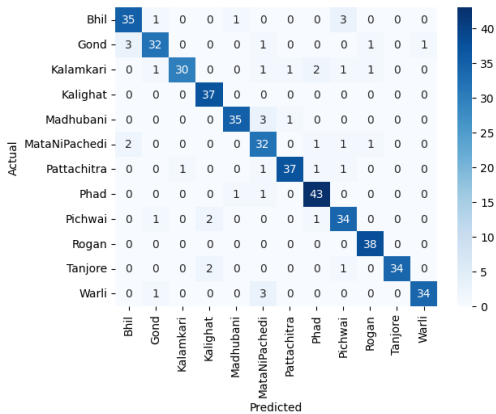

Fig. 5: Confusion Matrix for ResNet50

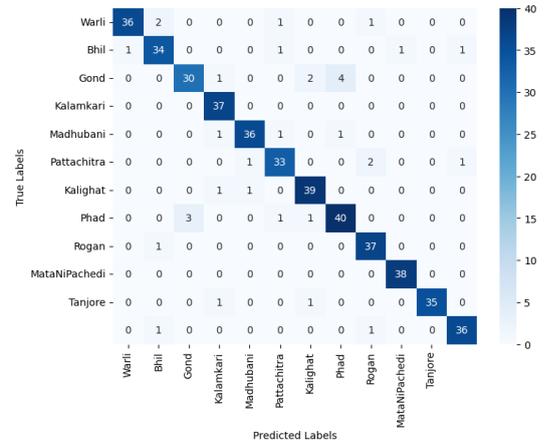

Fig. 6: Confusion Matrix for EfficietNetB0

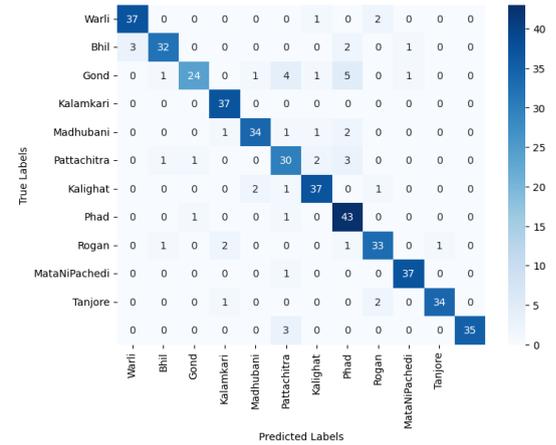

Fig. 6: Confusion Matrix for Inception V3

Table 4: Combined results (Accuracy) for ResNet50, EfficientNetB0, InceptionV3

| Art Form | InceptionV3 | ResNet50 | EffcientNetB0 |
|---|---|---|---|
| Bhil | 84.21 | 87.5 | 89.47 |
| Gond | 64.86 | 84.21 | 81.08 |
| Kalamkari | 100 | 81.08 | 100 |
| Kalighat | 90.24 | 100 | 95.12 |
| Madhubani | 87.17 | 89.74 | 92.31 |
| Warli | 92.5 | 89.47 | 94.73 |
| Phad | 95.55 | 95.55 | 88.88 |
| Pichwai | 92.10 | 89.47 | 94.73 |
| MataNiPachedi | 97.37 | 86.49 | 100 |
| Pattachitra | 81.08 | 90.24 | 89.19 |
| Rogan | 86.84 | 100 | 97.36 |
| Tanjore | 91.89 | 91.89 | 94.59 |

To evaluate the performance for multi-label classification for tagging, a new metric called 'mean Average Precision (mAP)' is used because of its efficient handling of imbalanced classes, and threshold independence. Further, mAP considers both Precision and Recall into a single metric offering insights into the precision-recall trade-off for each class independently. Table 5 shows the mAP scores obtained corresponding to all three CNN-based models.

Table 5: Performance metrics on Tagging

| CNN Model | Training mAP | Validation mAP |
|---|---|---|
| InceptionV3 | 86.82 | 83.05 |
| ResNet50 | 97.53 | 81.13 |
| EfficientNetB0 | 90.33 | **84.15** |

**Hyperparameter tuning:** To optimize the performance of the models in both multi-class and multi-label settings, effective hyperparameter tuning is performed, which played a crucial role in fine-tuning model parameters to achieve optimal results. In the experimentation, various crucial hyperparameters pertinent to training deep neural networks have been explored, including learning rate, batch-size, and learning-rate reduction factor. Specifically, learning rates were sampled from the available options of 0.001 and 0.0001, while batch-size varied across 32, 64, and 128. For the learning rate reduction factor, values of 0.2 and 0.5 were used. This enhances convergence and stability during training. Subsequently, for RandomForest, the experimentation was done with hyperparameters including the number of estimators (*n_estimators*) and the maximum depth of each tree (*max_depth*), with values (100, 200, 400, 800, 1000) and (10, 25, 35, 50), respectively. Additionally, the effect of the minimum number of samples required to split an internal node (*min_sample_split*) was explored across 2, 4, 8, and 16. Upon comprehensive evaluation, a learning rate of 0.001, batch size of 32, and a reduction factor of 0.5 yielded optimal performance of 96.22 on validation set and 92.04 on test set for EfficinetNetB0. For ensembling, the combination of 100 estimators, a maximum depth of 25, and a minimum sample split of 8, exhibited superior performance. Further details and results can be referred to the ablation table provided in the supplement[3].

## 5. CONCLUSION

This work proposes a dataset, *FolkTalent*, that consists of 2273 paintings (digital images) across 12 different folk-art forms. To facilitate comprehensive understanding, a comprehensive array of tags describing colors, attributes, objects, patterns, and their unique elements of saliency, are generated using GPT-4 and subsequently verified by an expert.

A novel ensemble approach has been proposed for the effective classification of Indian folk paintings on the *FolkTalent* dataset. The ensemble approach showed a notable change in validation and test accuracy on the dataset with a score of 94.00 and 91.83 percent respectively. Furthermore, with tagging, the proposed method showed a remarkable mean Average Precision (mAP) score of 84.15% on the validation dataset.

For reproducibility, the dataset and the code will be shared. The methodology employed generates an average of 30 image tags for each of the 12 painting styles, which have potential applications within the domain.

## 6. LIMITATIONS AND FUTURE SCOPE

The proposed work is currently limited to twelve types of Indian folk paintings only, namely Warli, Bhil, Gond, Kalamkari, Madhubani, Phad, Rogan, Tanjore, Mata Ni Pachedi, Kalighat, Pattachitra, and Pichwai. In the proposed dataset, each class contains an average of 190 paintings, which may affect developing standalone CNN based deep-neural models for tasks like image classification or tagging. However, given the size, fine-tuning (domain adaptation) of a pre-trained model is feasible as proposed.

The work can be extended to other Indian folk-art forms like Pithora, Chittara, Cheriyal, Sohrai, Manjusha, Thangka, etc. with an increased per class image count. In addition to tagging, the work can also be adapted for object detection from these paintings which can later facilitate the creation of detailed narratives associated with each painting. The generated tags can be incorporated together with textual description/caption in online folk painting catalogs that can significantly improve the user experience by improving efficiency and accuracy of search and retrieval functions. Additionally, the tags can be used to detect the authentic paintings based on the language/narrative (in the form of symbols, objects etc.) used for each art form, thereby contributing to the preservation of Indian folk arts.

---

[3] https://shorturl.at/iwCQW